%% file: aaai25.tex
\documentclass[letterpaper]{article} 
\usepackage{aaai25}  
\usepackage{times}  
\usepackage{helvet}  
\usepackage{courier}  
\usepackage[hyphens]{url}  
\usepackage{graphicx} 
\usepackage{natbib}  
\usepackage{caption} 
\usepackage{algorithm}
\usepackage{algorithmic}
\usepackage{multirow}
\usepackage{booktabs}
\usepackage{amsmath}
\usepackage{cleveref}
\usepackage{amssymb}
\usepackage{color}

\usepackage{newfloat}
\usepackage{listings}
\DeclareCaptionStyle{ruled}{labelfont=normalfont,labelsep=colon,strut=off} 
\floatstyle{ruled}
\newfloat{listing}{tb}{lst}{}
\floatname{listing}{Listing}
\title{Correspondence-Guided SfM-Free 3D Gaussian Splatting for NVS}
\author{
    Wei Sun\textsuperscript{\rm 1},
    Xiaosong Zhang\textsuperscript{\rm 2},
    Fang Wan\textsuperscript{\rm 1},
    Yanzhao Zhou\textsuperscript{\rm 1},
    Yuan Li\textsuperscript{\rm 1}\thanks{Corresponding author.},
    Qixiang Ye\textsuperscript{\rm 1},
    Jianbin Jiao\textsuperscript{\rm 1}
}
\affiliations{
    \textsuperscript{\rm 1}School of Electronic, Electrical and Communication
Engineering, University of Chinese Academy of Sciences, China\\

    \textsuperscript{\rm 2}Beijing Academy of Artificial Intelligence, China\\
    sunwei162@mails.ucas.ac.cn, xszhang@baai.ac.cn, 
    \{wanfang, zhouyanzhao, liyuan23, qxye, jiaojb\}@ucas.ac.cn
}

\usepackage{bibentry}

\begin{document}

\maketitle

\begin{abstract}

Novel View Synthesis (NVS) without Structure-from-Motion (SfM) pre-processed camera poses—referred to as SfM-free methods—is crucial for promoting rapid response capabilities and enhancing robustness against variable operating conditions.

Recent SfM-free methods have integrated pose optimization, designing end-to-end frameworks for joint camera pose estimation and NVS.
However, most existing works rely on per-pixel image loss functions, such as L2 loss. In SfM-free methods, inaccurate initial poses lead to misalignment issue, which, under the constraints of per-pixel image loss functions, results in excessive gradients, causing unstable optimization and poor convergence for NVS.

In this study, we propose a correspondence-guided SfM-free 3D Gaussian splatting for NVS.
We use correspondences between the target and the rendered result to achieve better pixel alignment, facilitating the optimization of relative poses between frames. 
We then apply the learned poses to optimize the entire scene.
Each 2D screen-space pixel is associated with its corresponding 3D Gaussians through approximated surface rendering to facilitate gradient back-propagation.
Experimental results underline the superior performance and time efficiency of the proposed approach compared to the state-of-the-art baselines.
\end{abstract}

\section{Introduction}
Novel-view synthesis serves as a fundamental objective within the realm of computer vision. The recent surge in NVS popularity is largely attributable to the success of Neural Radiance Fields (NeRFs)~\cite{mildenhall2021nerf} and 3D Gaussian Splatting (3DGS)~\cite{kerbl20233d}. However, these methods require densely captured views with accurately labeled camera poses, which is often not feasible in practical scenarios. Often, camera poses are obtained from SfM methods like COLMAP~\cite{schonberger2016structure} as a pre-processing step to NeRF and 3DGS, which is not only time-consuming but also prone to fail due to its sensitivity to feature extraction errors and difficulties in handling textureless or repetitive regions.

Recent studies~\cite{bian2023nope, lin2021barf, wang2021nerf, fu2023colmap, jiang2024construct} have focused on reducing the reliance on SfM by integrating pose estimation directly within the NVS framework. However, we would like to note that existing approaches typically rely on per-pixel image loss functions (such as L2 loss) from a pair of RGB images and compute per-pixel color derivatives with respect to desired scene parameters. The rendered result and the target do not sufficiently overlap or align because the camera pose is inaccurate at the initial stage of optimization. This problem is exacerbated when there is significant camera movement between consecutive views, at which point achieving perfect per-pixel alignment between the rendered result and the target becomes even more challenging. This misalignment issue, under the constraints of per-pixel image loss, often results in excessive gradients, leading to instability in the optimization process and difficulty in convergence.

To address this problem, we introduce a \textbf{C}orrespondence-\textbf{G}uided SfM-free \textbf{3D} \textbf{G}aussian \textbf{S}platting for NVS (CG-3DGS), a novel approach that integrates 2D correspondence detection~\cite{sun2021loftr, tang2022quadtree}, and computes derivatives on associated points instead of on a fixed grid of pixels. Specifically, we detect the 2d correspondence to find a pixel matching between rendered and target images and design a novel loss function based on the pixel matching. We then develop an approximated surface rendering pipeline for 3D Gaussians, which propagates disturbances from the 2D screen space to the parameters of the 3D Gaussians for differentiable scene optimization. Our derivatives are dense and could account for long-range object motions through the correspondence-based loss function, naturally leading to better robustness in optimization.

Inspired by but fundamentally distinct from CF-3DGS~\cite{fu2023colmap}, we construct a two-step optimization pipeline: (i) We initialize an auxiliary 3D Gaussian set given frame t with depth back-projection, and we sample the next nearby frame t+1. Our goal is to learn an affine transformation that can transform the 3D Gaussians in frame t to render the pixels in frame t+1. Correspondence-based loss function provides the gradients for optimizing the affine transformation, which is essentially optimizing the relative camera pose between frames t – 1 and t. This process continues iteratively until we obtain all the relative poses between frames 0, 1, ..., t. (ii). We initialize another 3D Gaussians set, where we perform scene optimization with all the frames and their corresponding learned camera poses.

This paper primarily contributes the following:

\begin{itemize}
\item We introduce the correspondence-guided SfM-free 3D Gaussian Splatting for NVS, minimizing the impact of pixel misalignment.

\item We integrate the 3DGS framework with effective correspondence supervision without time-consuming surface rendering, offering a unified differentiable pipeline for NVS without SfM pre-processing. %
\item Our method boosts time efficiency, and delivers superior results compared to the state-of-the-art methods. %
\end{itemize}

\section{Related Work}
\label{sec:related_work}

\subsection{Novel View Synthesis}Various 3D scene representations are utilized to produce realistic images from new viewpoints, including planes~\cite{horry1997tour,hoiem2005automatic}, meshes~\cite{hu2020worldsheet,Riegler2020FVS, riegler2021stable}, point clouds~\cite{xu2022point,zhang2022differentiable}, and multi-plane images~\cite{tucker2020single, zhou2018stereo, li2021mine}. NeRFs~\cite{mildenhall2021nerf} have recently become prominent for their superior photorealistic rendering capabilities, with numerous enhancements such as better anti-aliasing~\cite{barron2021mip, barron2022mip, barron2023zip, zhang2020nerf++} and improved reflectance~\cite{verbin2022ref, Attal_2023_CVPR}.

More recently, the use of point-cloud-based representations has surged due to its rendering efficiency~\cite{xu2022point, zhang2022differentiable, kerbl20233d, luiten2023dynamic, kopanas2022neural, yifan2019differentiable}. For example, Zhang~\cite{zhang2022differentiable} introduce a method to learn the per-point position and view-dependent appearance through a differentiable splat-based renderer initialized from object masks. Furthermore, 3DGS~\cite{kerbl20233d} facilitates real-time rendering of novel views using its explicit representation combined with a novel differential point-based splatting technique. Nevertheless, these methods typically depend on pre-computed camera parameters derived from SfM techniques~\cite{hartley2003multiple, schonberger2016structure, mur2015orb, taketomi2017visual}.

\subsection{SfM-Free Modeling for Novel View Synthesis}Initial efforts in SfM-free novel view synthesis include iNeRF~\cite{yen2021inerf}, which employs key-point matching to estimate camera poses. NeRFmm~\cite{wang2021nerf} introduces a method for joint optimization of camera pose and NeRF itself. Techniques such as those proposed in BARF~\cite{lin2021barf} focus on learning neural 3D representations and aligning camera frames using hierarchical positional encodings. The approach in Nope-NeRF~\cite{bian2023nope} incorporates monocular depth priors to simultaneously capture relative poses and synthesize new views.~\cite{meuleman2023progressively} uses a combination of pre-trained depth and optical-flow priors to refine blockwise NeRFs, which helps in the sequential recovery of camera poses.

In more generalizable settings, methods like SRT~\cite{srt22}, VideoAE~\cite{lai2021video}, RUST~\cite{sajjadi2023rust}, MonoNeRF~\cite{tian2023mononerf}, DBARF~\cite{chen2023dbarf}, and FlowCam~\cite{smith2023flowcam} aim to learn a scene representation from unposed videos using the implicit framework of NeRF. Despite these efforts, they often fail to achieve satisfactory view synthesis without specific scene optimization and share NeRF's original limitations, such as the inability to render explicit primitives in real time.

The inherent complexity of NeRF's implicit modeling often complicates the simultaneous optimization of scene and camera poses. Recent advancements like 3DGS, with its explicit point-based scene representation, facilitate real-time rendering and efficient optimization. New developments, such as those in CF-3DGS~\cite{fu2023colmap}, continue to push the limits of simultaneous scene and pose optimization, CF-3DGS~\cite{fu2023colmap} employs a progressive training strategy to reduce the cumulative noise associated with the pose optimization process. However, these methods consistently rely on per-pixel image loss, which always results in excessive gradients and unstable optimization when the optimization target and rendering result deviate significantly from perfect per-pixel alignment. This is a common issue in SfM-free scenarios due to the inaccurate initial pose estimation, which leads us to explore the integration of correspondence in the simultaneous scene and pose optimization.

\section{Method}
\label{sec:method}

In this paper, we leverage 3D Gaussians to reconstruct photo-realistic scenes from sequential frames of a video stream. 
Given a sequence of unposed images $\{ I_1, \dots, I_{K}\}$ with camera intrinsics,
our goal is to better reconstruct the complete scene via a joint optimization of the camera poses and the 3D representation (i.e. 3D Gaussians). We detail our method in the following sections, starting from a brief review of the representation and rendering process of 3D Gaussians~(Sec 3.1). Then, we propose a correspondence-guided pose optimization, a simple yet effective method to estimate the relative camera pose from each pair of nearby frames~(Sec 3.2). Finally, we briefly introduce how to reconstruct scenes using the estimated poses~(Sec 3.3).
\begin{figure*}[t]
\centering
\includegraphics[width = \linewidth]{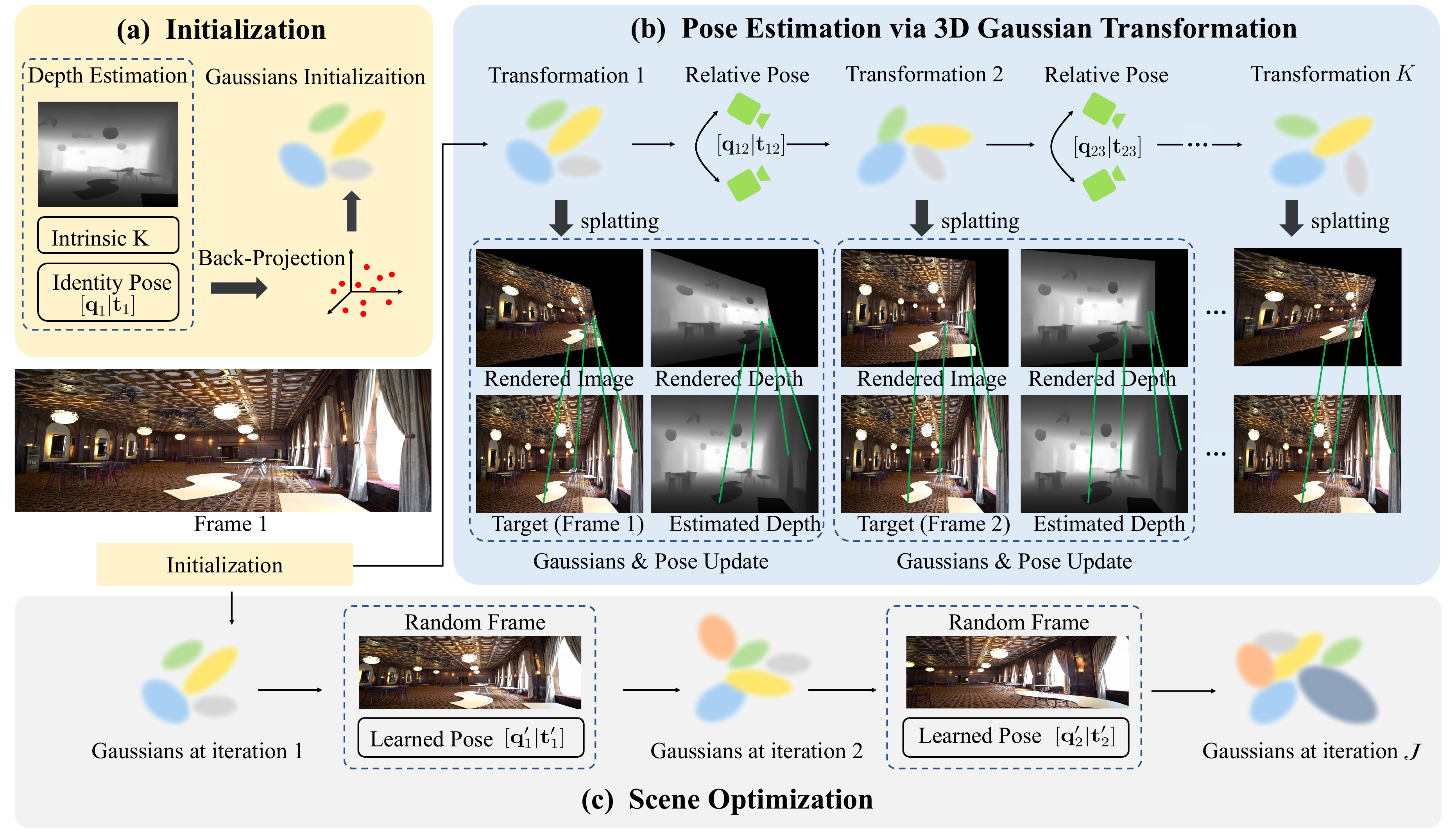}
\caption
{Overview of our CG-3DGS. (a) We utilize camera intrinsics and the identity pose to back-project depth estimate into a point cloud, initializing a set of 3D Gaussians. (b) These 3D Gaussians are used to simulate camera pose changes between adjacent frames through SE-3 transformations. First, we update the parameters of \( G_t \) based on the rendering results on frame \( t \), and use the SE-3 transformed \( G_t \) as \( G_{t+1} \) to render frame \( t+1 \). At this point, we freeze \( G_t \) and only update the parameters of the SE-3 transformation. This iterative process continues until the relative poses between all adjacent frames in a video sequence are estimated. The optimization is based on the correspondence between the rendered result and the ground truth. (c) After pose estimation, the same point cloud is also used to initialize a set of 3D Gaussians used for rendering the scene, and frames with estimated poses are randomly sampled for conventional training of the 3D Gaussians.}
\label{fig:overview}
\end{figure*}

\subsection{Revisiting 3D Gaussian Splatting}
3D Gaussian Splatting (3DGS) is a point-based novel view synthesis technique that uses 3D Gaussians to model the scene.  The Gaussian attributes are optimized based on a set of input training views denoted by ground truth images $\mathcal{I}_{gt} = \{I_i \in \mathbb{R}^{H \times W}\}_{i=1}^K$ and associated camera poses $\mathcal{P}_{gt} = \{W_i \in \mathbb{R}^{3 \times 4}\}_{i=1}^K$.
The Gaussian initialization is derived from a sparse point cloud created via SfM across the training views. To increase the number of Gaussians in areas where small-scale geometry is insufficiently reconstructed, a Gaussian densification process is periodically applied during training.

Each Gaussian $G_i$ in the scene is described by several parameters: its position $x_i \in \mathbb{R}^3$, scale $s_i \in \mathbb{R}^3$, rotation $r_i \in \mathbb{R}^4$, base color $c_i \in \mathbb{R}^3$, view-dependent spherical harmonics $h_i \in \mathbb{R}^{15\times 3}$, and opacity $\alpha_i \in \mathbb{R}$. Collectively, these parameters are grouped as:
\begin{equation}
\mathcal{G} = \{G_i = \{x_i, s_i, r_i, c_i, h_i, \alpha_i\}\}_{i=1}^N,
\end{equation}
where $N$ denotes the total count of Gaussians.

During synthesis, scaling and rotation parameters are translated into matrices $S_i$ and $R_i$. The Gaussian $G_i$ is spatially characterized in the 3D scene by its center point, or mean position, $x_i$ and a decomposable covariance matrix $\Sigma_i$:
\begin{equation}
    G_i(x_i) = e^{-\frac{1}{2}x_i^T\Sigma^{-1}_i x_i}, \Sigma_i = R_iS_iS_i^TR_i^T.
\end{equation}

To facilitate the differentiation of 3D Gaussian rendering, the Gaussian projection process is applied from a specific camera pose $W$, approximating the splatting of a 3D Gaussian onto the 2D image plane:
\begin{equation}
    \Sigma^{\text{2D}} = J W \Sigma W^{\top} J^{\top}
\end{equation}
where $J$ represents the Jacobian of the affine approximation of the projective transformation.

For each pixel, the final rendered color and depth can be formulated as the alpha-blending of $N$ ordered Gaussians that overlap the pixel:
\begin{equation}
\label{eq:gauss}
\begin{aligned}
    \hat{C} &= \sum_i^N \mathbf{c}_i \alpha_i \prod_j^{i-1} (1-\alpha_j), \\
    \hat{D} &= \sum_i^N \mathbf{d}_i \alpha_i \prod_{j}^{i-1}(1-\alpha_j),
\end{aligned}
\end{equation}
with $\mathbf{c}_i$, $\alpha_i$, and $\mathbf{d}_i$ representing the color, opacity, and depth of the Gaussians, respectively.

The optimization of the 3DGS model relies on minimizing a composite loss function using stochastic gradient descent:

\begin{equation}
    L(G | W, I) = ||\hat{I} - I||_1 + L_{SSIM}(\hat{I}, I),
\end{equation}
where $\hat{I}$ is the rendering result and $I$ is the ground truth image. The overall loss combines L1 loss for residual minimization and SSIM loss for structural similarity.

\subsection{Correspondence-guided Pose Optimization}\label{method:pose}

\subsubsection{Initialization from Monocular Depth.}
As shown in Fig.~\ref{fig:overview} (a), for the initial frame $I_1$, which is at timestep $1$, we apply a standard monocular depth network to produce a depth map, represented as $D_1$. We then construct the point cloud $P$ by back-projecting the depth map $D_1$ using the default identity camera pose (orthogonal projection) and camera intrinsics, and use $P$ to initialize 3D Gaussians instead of relying on SfM-derived points. Following this initialization, we optimize a set of 3D Gaussians $G_1$, adjusting all attributes to reduce the correspondence-based loss between the rendered image and the ground truth $I_1$,
\begin{equation}
    {G_1}^{\star} = \arg \min_{c_1, r_1, s_1, \alpha_1} \mathcal{L}_{cor}(\mathcal{R}(G_1), I_1),
\end{equation}
where $\mathcal{R}$ denotes the rendering operation of 3DGS. The correspondence-based loss $\mathcal{L}_{cor}$ is detailed in Sec.~3.2.3.

\subsubsection{Pose Estimation via 3D Gaussians Transformation.}
The problem of camera pose estimation is addressed by predicting the transformation of 3D Gaussians as discussed in CF-3DGS~\cite{fu2023colmap}. Starting with the Gaussian center's position $\mu$, we project it into the 2D camera plane with camera pose $W$ as $\mu_{2D} = K\frac{W \mu}{(W \mu)_z}$. Hence, estimating the camera pose effectively involves determining the transformations of these 3D Gaussians.

For the relative camera pose estimation, we apply a learnable SE-3 affine transformation $T_t$ to the pretrained 3D Gaussians ${G_t}^*$, transforming it into frame $t+1$, represented as $G_{t+1}= T_t \odot G_t$. This transformation $T_t$ is refined by minimizing the photometric loss between the rendered images and the subsequent frame $I_{t+1}$:
\begin{equation}
    {T_t}^* = \arg \min_{T_t} \mathcal{L}_{cor}(\mathcal{R}(T_t \odot G_t), I_{t+1}),
\end{equation}
During this optimization phase, we preserve the attributes of the pretrained 3D Gaussians ${G_t}^*$ unchanged to distinctly separate the effects of camera motion from changes such as deformation, densification, pruning, or self-rotation of the Gaussians. 
The transformation matrix $T$, comprising quaternion rotations $\mathbf{q} \in \mathfrak{so}(3)$ and translation vectors $\mathbf{t} \in \mathbb{R}^3$, facilitates the estimation of relative camera poses between consecutive frames. After this, we have estimated the relative camera pose between frames $I_t$ and $I_{t+1}$. As the next frame $I_{t+2}$ becomes available, this process is repeated: we optimize the 3D Gaussians to obtain \( {G^*_{t+1}} \), similar to what is described at the end of Sec.~3.2.1; we then optimize the relative pose between \( I_{t+1} \) and \( I_{t+2} \), and could subsequently infer the relative pose between \( I_{t} \) and \( I_{t+2} \).

\subsubsection{Correspondence-based Loss}
\label{method:correspondence}
We utilize off-the-shelf detectors \cite{tang2022quadtree, sun2021loftr} to establish the 2D correspondences between ground truth image $I$ and rendered result $\hat{I}(W)$ for the pose optimization. The 2D screen-space coordinates in $I$ are represented as $\mathcal{K}=\{\mathbf{k}^{(1)}, \mathbf{k}^{(2)}, ..., \mathbf{k}^{(M)}\}$, where $M$ represents the total number of points. Correspondingly, the 2D screen-space coordinates in $\hat{I}(W)$ are $\mathcal{K}'=\{\mathbf{k}^{'(1)}, \mathbf{k}^{'(2)}, ..., \mathbf{k}^{'(M)}\}$. The optimization objective is to align $\mathbf{k}$ with $\mathbf{k}'$, visualized in Fig.~\ref{fig:overview}~(b).

To enable gradient back-propagation from the matching of $\mathbf{k}$ and $\mathbf{k}'$ to the 3D Gaussians shaping the surface, we employ a differentiable approximate surface renderer, described in Sec.~3.2.4, to render the screen-space coordinates at $\mathbf{k}^{'(i)}, i=1,2,...,M$ as $q(\mathbf{k}^{'(i)})$.
The resulting loss function is expressed as:
\begin{equation}
    \mathcal{L}_\text{cor-rgb} = \sum_{i=1}^M || q(\mathbf{k}^{'(i)}) - \mathbf{k}^{(i)} ||_1.
\end{equation}
Notably, $q(\mathbf{k}^{'(i)})$ numerically matches $\mathbf{k}^{'(i)}$, yet it creates a pathway for gradients to flow back to the underlying 3D representation without altering the original 3DGS.

Moreover, incorporating short-range relations through pixel-wise supervision can assist in stabilizing the optimization process. 
This loss is formulated as:
\begin{equation}
    \mathcal{L}_\text{pix-rgb} = || I - \hat{I}(W) ||_1.
\end{equation}

Furthermore, the depth matching process involves equating the monocular depth at $\mathbf{k}$, denoted as $d(\mathbf{k})$, with the rendered depth at $\mathbf{k}^{'}$, denoted as $\hat{d}(\mathbf{k}^{'})$. 
The corresponding loss term is defined as:
\begin{equation}
    \mathcal{L}_\text{cor-depth} = \sum_{i=1}^M || \hat{d}(\mathbf{k}^{'(i)}) - d(\mathbf{k}^{(i)}) ||_1.
\end{equation}

The correspondence-based loss consolidates these components:
\begin{equation}
\label{eqn:optimization-objective}
    \mathcal{L}_\text{cor} = \lambda_1 \mathcal{L}_\text{cor-rgb} + \lambda_2 \mathcal{L}_\text{pix-rgb} + \lambda_3 \mathcal{L}_\text{cor-depth}, 
\end{equation}
where $\lambda_1 = 10, \lambda_2 = 1, \lambda_3 = 1$.

\begin{figure}[t]
\centering
\includegraphics[width=\linewidth]{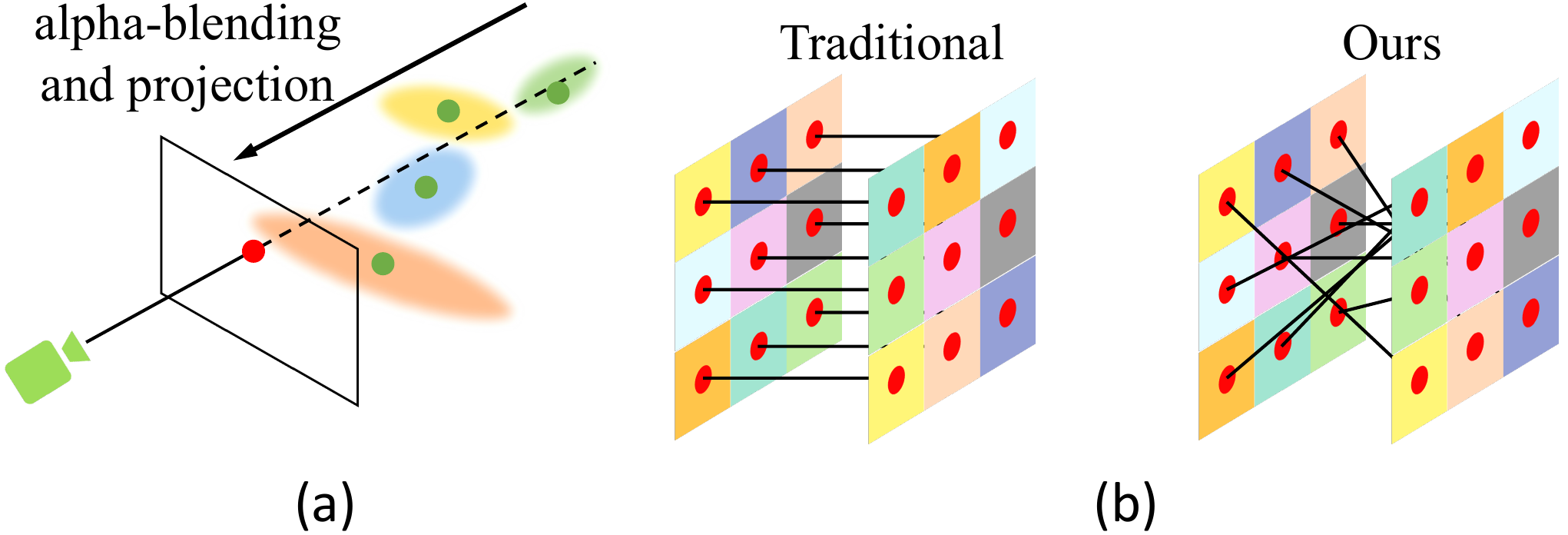}
\caption{(a) By performing alpha-blending on the center coordinates of the 3D Gaussians, an approximate 3D surface point is generated and projected onto the 2D screen. (b) The comparison between traditional methods and our method. The fundamental difference is the technique used for aligning pixels.}
\label{fig:fig3}
\end{figure}

\input{tables/sota_nvs_tanks}

\subsubsection{Approximated Surface Rendering}
\label{methodology-differentiable-approximate-surface-rendering}
Our aim in correspondence-based optimization is to transmit gradient information from a 2D screen-space location to its associated 3D surface location. Essentially, we seek to link disturbances at a 2D screen-space location with those at its 3D surface counterpart. 

Given the volumetric nature of the 3D Gaussian representation, explicit surfaces are not present. However, reconstructing an explicit surface is extremely time-consuming~\cite{park2019deepsdf}, and modifying the rendering logic of the 3D Gaussian to obtain surfaces would also significantly increase the training duration~\cite{jiang2024construct}. Instead, according to previous studies \cite{keetha2024splatam, chung2024depth, fu2023colmap, yan2024gs}, the depth of an expected 3D surface point $D(\mathbf{k})$ relative to a 2D screen-space point $\mathbf{k}$ is computed as follows:
\begin{equation}
\label{equation-native-depth}
\begin{aligned}
    D(\mathbf{k}) &= \sum_{i} d_i \alpha_i(\mathbf{k}) \prod_{j=1}^{i-1} (1 - \alpha_j(\mathbf{k})),
\end{aligned}
\end{equation}
where $d_i$ indicates the $z$-axis position of the Gaussian centers within the camera coordinate system, and $\alpha_i$ and $\alpha_j$ represent the alpha-blending coefficients for the $i$\textsuperscript{th} and $j$\textsuperscript{th} Gaussian, respectively.

As illustrated in Fig.~\ref{fig:fig3}, the corresponding expected 3D surface point $\Psi(\mathbf{k})$ at $\mathbf{k}$ could then be defined by:
\begin{equation}
\begin{aligned}
    \Psi(\mathbf{k}) &= \sum_{i} {\boldsymbol{\mu}}_i \alpha_i(\mathbf{k}) \prod_{j=1}^{i-1} (1 - \alpha_j(\mathbf{k})), 
\end{aligned}
\end{equation}
where ${\boldsymbol{\mu}}_i\in\mathbb{R}^3$ represents the center position of the $i^\mathrm{th}$ Gaussian. This formula provides an approximation of the 3D surface point without relying on time-consuming surface reconstruction method like signal distance function (SDF)~\cite{park2019deepsdf}.

\subsubsection{Correspondence Cache Mechanism.}
In our pose optimization process, correspondences between the rendered views and the reference views are stored in a cache. By reusing these correspondences in subsequent iterations, we achieve a significant reduction in computation time without significantly degrading performance, as adjacent frames in continuous video tend to exhibit similar features and poses. Concretely, rather than recalculating correspondence points for each image pair in every iteration, we strategically update these correspondences every \(H\) iterations—where \(H\) is empirically set to 50.

\subsection{Scene Optimization}\label{sec: gaussian_optimize}
Following the camera pose optimization, we proceed to optimize a new set of 3D Gaussians that ultimately represent the scene. Similarly to the pose optimization phase, we start by generating a set of initialized 3D Gaussians using the depth estimation from the first frame $I_1$. Here, we keep the camera pose fixed and focus solely on minimizing the photometric loss as in the vanilla 3DGS~\cite{kerbl20233d}. During the optimization, we randomly sample frames from the training set of the scene and utilize the associated optimized poses for training, as shown in Fig.~\ref{fig:overview} (c).

\input{tables/sota_pose_tanks}

\begin{figure*}[tp]
    \centering
    
    \includegraphics[width=1.0\linewidth]{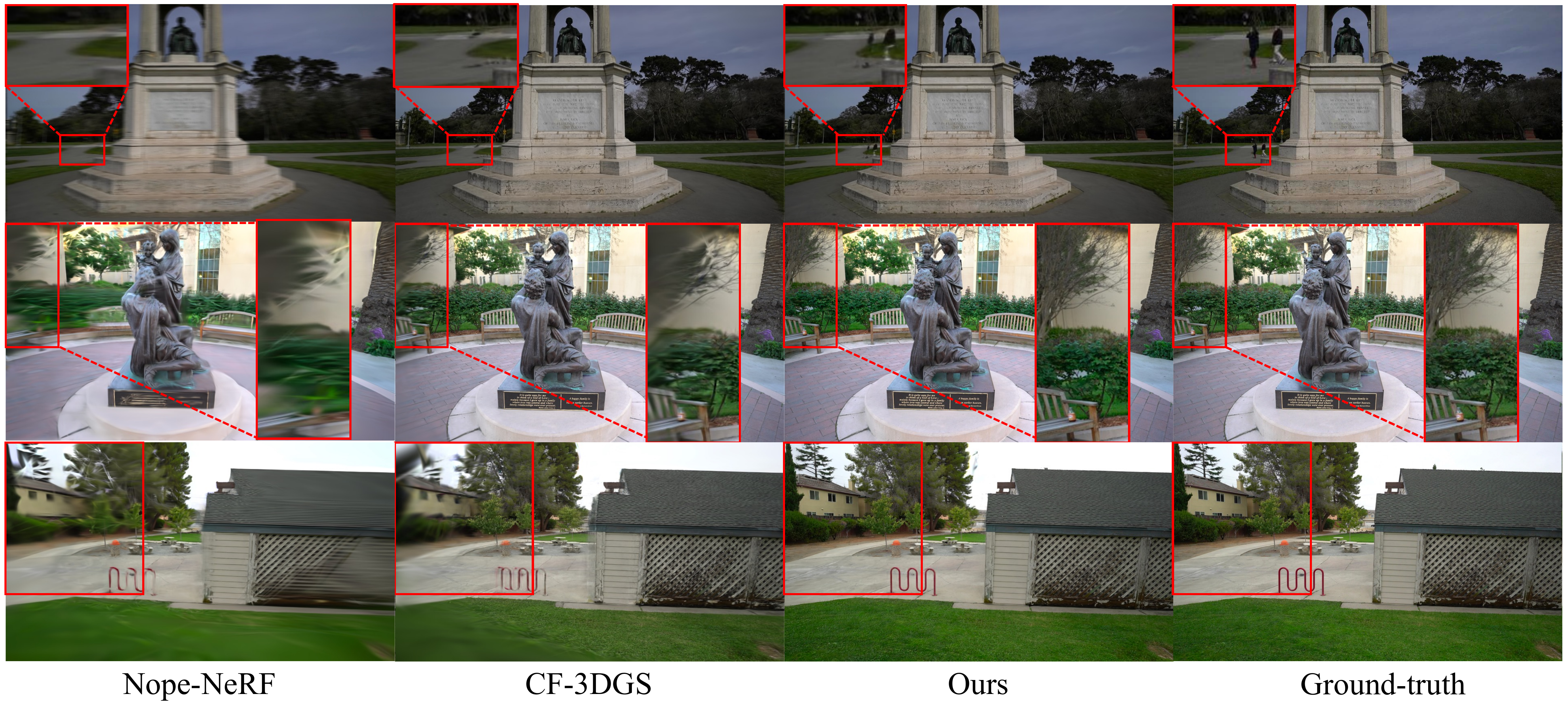}
    \caption{\textbf{Qualitative comparison for novel view synthesis on Tanks and Temples.} Our approach produces more realistic rendering results than other baselines. \textbf{Better viewed when zoomed in.}}
    \label{fig:nvs}
\end{figure*}

\section{Experiments}
\label{sec:exp}

\subsection{Experimental Setup}

\noindent\textbf{Datasets.}
We conduct extensive experiments on several real-world datasets, including Tanks and Temples~\cite{Knapitsch2017} and CO3D-V2~\cite{reizenstein2021common}. \textbf{Tanks and Temples:} Following the methodology in Nope-NeRF~\cite{bian2023nope}, we assess the quality of novel view synthesis and the accuracy of pose estimation across eight diverse scenes that encompass both indoor and outdoor environments. We select seven images from each 8-frame sequence for training and evaluate the novel view synthesis on the remaining frame. Camera poses are estimated and assessed on all training images following alignment according to Umeyama's method~\cite{umeyama1991least}. \textbf{CO3D-V2:} This dataset comprises thousands of object-centric videos, maintaining view of the full object while the camera moves in a complete circle around it. Deriving camera poses from CO3D videos is more challenging compared to Tanks and Temples due to the large and complex camera movements. We randomly select 
four scenes from different object categories and follow the same protocol as CF-3DGS~\cite{fu2023colmap} to divide the training and testing sets.

\noindent\textbf{Metrics.}
We assess the performance of novel view synthesis and camera pose estimation tasks. For the former, we evaluate using standard metrics such as Peak Signal-to-Noise Ratio (PSNR), Structural Similarity Index (SSIM)~\cite{wang2004image}, and Learned Perceptual Image Patch Similarity (LPIPS)~\cite{zhang2018unreasonable}. For the latter, we utilize established visual odometry metrics such as Absolute Trajectory Error (ATE) and Relative Pose Error (RPE).

\input{tables/sota_nvs_co3d}
\input{tables/sota_pose_co3d}
\input{tables/abs_grow}

\noindent\textbf{Implementation Details.} Our implementation leverages the PyTorch framework~\cite{paszke2017automatic} and adheres to the optimization parameters specified in 3DGS~\cite{kerbl20233d}, unless noted otherwise. Importantly, we continuously adjust the opacity throughout the training process to effectively limit the unchecked growth of Gaussian components caused by inaccuracies in pose estimation. For the Tanks and Temples and CO3D V2 datasets, the off-the-shelf monocular depth networks used are DPT~\cite{ranftl2021vision} and Zoe~\cite{bhat2023zoedepth}, respectively. The initial learning rate is set at $10^{-5}$ and is progressively reduced to $10^{-6}$ until the model converges. All experiments are performed on a single RTX 3090 GPU.

\subsection{Comparing with SfM-Free Methods}
In this subsection, we compare our method with several baselines including CF-3DGS~\cite{fu2023colmap}, Nope-NeRF~\cite{bian2023nope}, BARF~\cite{lin2021barf} and  NeRFmm~\cite{wang2021nerf} on both novel view synthesis and camera pose estimation.

\subsubsection{Novel View Synthesis.} In contrast to conventional approaches where camera poses for testing views are provided, we need to additionally ascertain the camera poses of test views. We utilize the same protocol as outlined in CF-3DGS to optimize the camera poses for these testing views. This identical procedure is applied across all baseline models to maintain a consistent basis for comparison.

We present the comparative analysis on the Tanks and Temples dataset in Table~\ref{table:nvs}. Our approach consistently surpasses competing methods across all evaluated metrics. Remarkably, our direct training strategy achieves superior PSNR values even compared to CF-3DGS, which leverages carefully designed progressive 3D Gaussians training strategy, with a notable increase of 3.5 points in the Family scene.

Qualitative results are shown in Fig.~\ref{fig:nvs}. The images generated using our method are distinctly sharper and could retain small objects within the scene, such as the walking person in the first scene shown in Fig.~\ref{fig:nvs}, which correlates with the significantly improved scores for SSIM and LPIPS, as detailed in Table~\ref{table:nvs}.

\subsubsection{Camera Pose Estimation.} The learned camera poses are post-processed by Procrustes analysis as in~\cite{lin2021barf, bian2023nope} and compared with the ground-truth poses of training views. The quantitative results of camera pose estimation are summarized in Table~\ref{table:pose}. Our approach achieves comparable performance with the current state-of-the-art results. 
We hypothesize that the relatively poorer performance in terms of RPE$_r$ may be attributed to relying solely on photometric loss for relative pose estimation. In contrast, Nope-NeRF incorporates additional constraints on relative poses beyond photometric loss, including the chamfer distance between two point clouds. As indicated in~\cite{bian2023nope}, omitting the point cloud loss leads to a significant decrease in pose accuracy.

\subsection{Performance in Complex Camera Motions}
While the camera motions involved in the Tanks and Temples dataset are relatively minor, we extend the validation of our method's robustness to the CO3D videos, which feature more intricate and demanding camera movements. 

As shown in Table~\ref{table:nvs_co3d} and Table~\ref{table:pose_co3d}, our approach not only excels in novel view synthesis but also clearly surpasses CF-3DGS in pose estimation, reinforcing the findings from the Tanks and Temples experiments and underscoring the precision and robustness of our proposed method in scenarios characterized by complex camera motions.

\input{tables/abs_colmap}

\subsection{Ablation Study}

\subsubsection{Effectiveness of Correspondence.} We assess the impact of correspondence-guided optimization by substituting it for traditional pixel-wise supervision. Performance metrics for novel view synthesis and camera pose estimation with and without correspondence-guided optimization are detailed in Table~\ref{table:abs_grow}. Our observations confirm that correspondence plays a crucial role in enhancing both novel view synthesis and pose estimation accuracy. In the absence of correspondence, inaccurate initial poses lead to significant deviations between the screen space coordinates of objects in the rendered images and those in the GT images, resulting in poor gradients quality and unstable optimization of the 3D Gaussians model.

\subsubsection{Comparison with 3DGS with SfM Poses.} Our analysis extends to comparing the quality of novel view synthesis of our method with that of the conventional 3DGS model~\cite{kerbl20233d}, which utilizes poses derived from SfM technique on the Tanks and Temples dataset. As shown in Table~\ref{table:abs_colmap}, our integrated optimization framework delivers performance on par with the 3DGS model that incorporates SfM-derived poses. In scenes where SfM pose estimation is challenging, there is a significant improvement in performance, as observed in the Horse scene.

\section{Conclusion}
We introduce a novel correspondence-guided SfM-free 3D Gaussian splatting for NVS method that enhances novel-view synthesis by avoiding SfM pre-processing. Our approach effectively optimizes relative poses between frames through correspondence estimation and achieves a differentiable pipeline using our proposed approximated surface rendering technique. Experimental results confirm the superiority of our method in terms of quality and efficiency.

\bibliography{aaai25}

\end{document}

%% file: tables/sota_nvs_tanks.tex
\begin{table*}[ht]
\setlength{\tabcolsep}{2.5pt}
\centering
\footnotesize
\begin{tabular}{cccccccccccccccccccccc}
\hline
& \multirow{2}{*}{scenes} &  &
\multicolumn{3}{c}{Ours} &  &
\multicolumn{3}{c}{CF-3DGS} &  & \multicolumn{3}{c}{Nope-NeRF}  &  & \multicolumn{3}{c}{BARF} &  & \multicolumn{3}{c}{NeRFmm} \\ 
\cline{4-6} \cline{8-10} \cline{12-14} \cline{16-18} \cline{20-22} &  &  & PSNR $\uparrow$  & SSIM $\uparrow$  & LPIPS $\downarrow$   &  & PSNR  & SSIM  & LPIPS &  & PSNR & SSIM  & LPIPS  &  & PSNR   & SSIM & LPIPS  &  & PSNR  & SSIM  & LPIPS  \\ \hline
  & Church &  & \textbf{32.14} & \textbf{0.96} & \textbf{0.08}                 &  & 30.23 & 0.93 & 0.11  & & 25.17 & 0.73 & 0.39 &  & 23.17 & 0.62          & 0.52  &  & 21.64   & 0.58    & 0.54            \\
  & Barn &  & \textbf{33.19} & \textbf{0.94} & \textbf{0.07}                   &  & 31.23 & 0.90 & 0.10  & & 26.35 & 0.69 & 0.44 &  & 25.28 & 0.64          & 0.48  &  & 23.21   & 0.61    & 0.53             \\
  & Museum  &  & \textbf{31.62} & \textbf{0.94} & \textbf{0.08}                &  & 29.91 & 0.91 & 0.11 & & 26.77 & 0.76 & 0.35 &  & 23.58 & 0.61          & 0.55  &  & 22.37   & 0.61    & 0.53            \\
  & Family &  & \textbf{34.80} & \textbf{0.97} & \textbf{0.04}                 &  & 31.27 & 0.94 & 0.07 & & 26.01 & 0.74 & 0.41 &  & 23.04 & 0.61          & 0.56  &  & 23.04   & 0.58    & 0.56             \\
  & Horse &  & \textbf{35.45} & \textbf{0.97} & \textbf{0.04}                  &  & 33.94 & 0.96 & 0.05 & & 27.64 & 0.84 & 0.26 &  & 24.09 & 0.72          & 0.41  &  & 23.12   & 0.70    & 0.43             \\
  & Ballroom  &  & \textbf{33.91} & \textbf{0.97} & \textbf{0.04}              &  & 32.47 & 0.96 & 0.07 & & 25.33 & 0.72 & 0.38 &  & 20.66 & 0.50          & 0.60  &  & 20.03   & 0.48    & 0.57             \\
  & Francis  &  & \textbf{33.80} & \textbf{0.92} & \textbf{0.13}               &  & 32.72 & 0.91 & 0.14 & & 29.48 & 0.80 & 0.38 &  & 25.85 & 0.69          & 0.57  &  & 25.40   & 00.69   & 0.52             \\
  & Ignatius &  & \textbf{31.14} & \textbf{0.94} & \textbf{0.06}               &  & 28.43 & 0.90 & 0.09 & & 23.96 & 0.61 & 0.47 &  & 21.78 & 0.47          & 0.60  &  & 21.16   & 0.45    & 0.60             \\ \hline
  & mean  &  & \textbf{33.26} & \textbf{0.95} & \textbf{0.07}                  &  & 31.28 & 0.93 & 0.09 & & 26.34 & 0.74 & 0.39 &  & 23.42 & 0.61          & 0.54  &  & 22.50   & 0.59    & 0.54             \\ \hline
\end{tabular}
\caption{\textbf{Novel view synthesis results on Tanks and Temples}. Each baseline method is trained with its public code under the original settings and evaluated with the same evaluation protocol. The best results are highlighted in bold.}
\label{table:nvs}
\end{table*}

%% file: tables/sota_pose_tanks.tex
\begin{table*}[t]
\setlength{\tabcolsep}{3.4pt}
\centering
\footnotesize
\begin{tabular}{cccccccccccccccccccccc} 
\hline
\multirow{2}{*}{} & \multirow{2}{*}{scenes} &  & \multicolumn{3}{c}{Ours} &  & \multicolumn{3}{c}{CF-3DGS} &  & \multicolumn{3}{c}{Nope-NeRF} &  &  \multicolumn{3}{c}{BARF} &  & \multicolumn{3}{c}{NeRFmm}     \\ 
\cline{4-6} \cline{8-10} \cline{12-14} \cline{16-18} \cline{20-22} &  &  & $\text{RPE}_t \downarrow$ & $\text{RPE}_r \downarrow$ & ATE$ \downarrow$ &  & $\text{RPE}_t $ & $\text{RPE}_r $ & ATE &  &  $\text{RPE}_t $ & $\text{RPE}_r $ & ATE &  & $\text{RPE}_t$ & $\text{RPE}_r$ & ATE   &  & $\text{RPE}_t$ & $\text{RPE}_r$ & ATE    \\ 
\hline

& Church   &  & 0.006          & 0.016          & 0.001               &  & 0.008 & 0.018 & 0.002 &  & 0.034 & 0.008  & 0.008  &  & 0.114 &0.038 & 0.052 &  & 0.626          & 0.127          & 0.065   \\
& Barn   &  & 0.029          & 0.030         & 0.002                 &  & 0.034 & 0.034 & 0.003  &  & 0.046 & 0.032  &  0.004 &  & 0.314 &0.265 & 0.050  &  & 1.629          & 0.494          & 0.159   \\
& Museum  &  & 0.047          & 0.203          & 0.004                &  & 0.052 & 0.215 & 0.005  &  &  0.207 & 0.202 & 0.020   &  & 3.442 &1.128 & 0.263   &  & 4.134          & 1.051          & 0.346   \\
& Family   &  & 0.024 & 0.020 & 0.001               &  & 0.022 & 0.024 & 0.002 & & 0.047 & 0.015 & 0.001  &  & 1.371 &0.591 & 0.115  &  & 2.743 & 0.537 & 0.120   \\
& Horse    &  & 0.109          & 0.053          & 0.003                &  & 0.112 & 0.057 & 0.003 & & 0.179 & 0.017 & 0.003   &  & 1.333 &0.394 & 0.014  &  & 1.349          & 0.434          & 0.018  \\
& Ballroom    &  & 0.033          & 0.020          & 0.003            &  & 0.037 & 0.024 & 0.003 & & 0.041  & 0.018 & 0.002   &  & 0.531 &0.228 & 0.018  &  & 0.449          & 0.177          & 0.031   \\
& Francis    &  & 0.026          & 0.147          & 0.005             &  & 0.029 & 0.154 & 0.006 &  &  0.057 & 0.009  & 0.005   &  & 1.321 &0.558 & 0.082  &  & 1.647          & 0.618          & 0.207   \\
& Ignatius  &  & 0.027          & 0.012         & 0.003              &  & 0.033 & 0.032 & 0.005 & &  0.026  & 0.005  & 0.002   &  & 0.736 &0.324 & 0.029  &  & 1.302          & 0.379          & 0.041   \\ \hline
& mean &  & \textbf{0.037} &0.063 & \textbf{0.003}                   &  & 0.041 & 0.069 & 0.004 & & 0.080  & \textbf{0.038} & 0.006  &  & 1.046 &0.441 & 0.078   &  & 1.735 &0.477 & 0.123   \\

\hline
\end{tabular}
\caption{\textbf{Pose accuracy on Tanks and Temples}. 
Note that we use COLMAP poses in Tanks and Temples as the ``ground truth". The unit of $\text{RPE}_r$ is in degrees, ATE is in the ground truth scale and $\text{RPE}_t$ is scaled by 100. The best results of means are highlighted in bold.
}
\vspace{-4mm}
\label{table:pose}
\end{table*}

%% file: tables/sota_nvs_co3d.tex
\begin{table*}[ht]
\centering
\setlength{\tabcolsep}{3.7pt}
\footnotesize
\begin{tabular}{ccccccccccccccccccccccc}
\toprule
\multirow{2}{*}{Method}  & \multirow{2}{*}{Times $\downarrow$} & & \multicolumn{3}{c}{46\_2587\_7531} & & \multicolumn{3}{c}{407\_54965\_106262} & & \multicolumn{3}{c}{429\_60388\_117059} & & \multicolumn{3}{c}{437\_62482\_122880}  \\ 
\cline{4-6} \cline{8-10} \cline{12-14} \cline{16-18} &  &  & PSNR $\uparrow$  & SSIM $\uparrow$  & LPIPS $\downarrow$   &  & PSNR  & SSIM  & LPIPS &  & PSNR & SSIM  & LPIPS  &  & PSNR   & SSIM & LPIPS  \\ \hline
Nope-NeRF & $\thicksim$30 h & &  25.3& 0.73& 0.46 & & 25.53& 0.83& 0.58 & &  22.19 &0.62 &0.56  & & 20.81 & 0.59 & 0.51 \\
CF-3DGS & $\thicksim$2 h & & 25.44 &0.80& 0.21 & & 27.80 &0.84 &0.35 & & 24.44 &0.68 &0.36 & & 22.95 & 0.66 & 0.41  \\
Ours & \textbf{$\thicksim$1.5 h} & & \textbf{26.43} & \textbf{0.85} & \textbf{0.15} &  &  \textbf{28.46} & \textbf{0.88} & \textbf{0.27} & & \textbf{25.72} & \textbf{0.74} & \textbf{0.29} & & \textbf{24.32}	& \textbf{0.69} & \textbf{0.32} \\
\bottomrule
\end{tabular}
\vspace{-2mm}
\caption{\textbf{Novel view synthesis results on CO3D V2}. Each baseline method is trained with its public code under the original settings and evaluated with the same evaluation protocol. The best results are highlighted in bold.}

\vspace{-0.05in}
\label{table:nvs_co3d}
\end{table*}

%% file: tables/sota_pose_co3d.tex
\begin{table*}[ht]
\centering
\setlength{\tabcolsep}{4.7pt}
\footnotesize
\begin{tabular}{ccccccccccccccccccccccc}
\toprule
\multirow{2}{*}{Method}  & \multirow{2}{*}{Times $\downarrow$} & & \multicolumn{3}{c}{46\_2587\_7531} & & \multicolumn{3}{c}{407\_54965\_106262} & & \multicolumn{3}{c}{429\_60388\_117059} & & \multicolumn{3}{c}{437
\_62482\_122880}\\ 
\cline{4-6} \cline{8-10} \cline{12-14} \cline{16-18} &  &  & $\text{RPE}_t \downarrow$ & $\text{RPE}_r \downarrow$ & ATE$ \downarrow$ &  & $\text{RPE}_t $ & $\text{RPE}_r $ & ATE &  &  $\text{RPE}_t $ & $\text{RPE}_r $ & ATE &  & $\text{RPE}_t$ & $\text{RPE}_r$ & ATE      \\ \hline
Nope-NeRF & $\thicksim$30 h & & 0.426 & 4.226 & 0.023 & &  0.553& 4.685 &0.057  & & 0.398 &2.914 &0.055  & & 0.591 & 2.014 & 0.041 \\
CF-3DGS & $\thicksim$2 h & & 0.095 & 0.447 & 0.009 &  &  0.31 &0.243 &0.008 & & 0.134& 0.542 &0.018 & & 0.252	& 0.493 & 0.018 \\
Ours & \textbf{$\thicksim$1.5 h} & & \textbf{0.041} & \textbf{0.274} & \textbf{0.005} &  &  \textbf{0.14} & \textbf{0.182} & \textbf{0.003} & & \textbf{0.092} & \textbf{0.239} & \textbf{0.008} & & \textbf{0.116}	& \textbf{0.284} & \textbf{0.009} \\
\bottomrule
\end{tabular}

\vspace{-2mm}
\caption{\textbf{Pose accuracy on CO3D V2}. Note that the camera poses provided by CO3D as the ``ground truth". The unit of $\text{RPE}_r$ is in degrees, ATE is in the ground truth scale and $\text{RPE}_t$ is scaled by 100. The best results are highlighted in bold.}
\vspace{-0.05in}
\label{table:pose_co3d}
\end{table*}

%% file: tables/abs_grow.tex
\begin{table}[tp]
\setlength{\tabcolsep}{1.3pt}
\footnotesize
\begin{tabular}{ccccccccccccccccccccccc}
\toprule
\multirow{2}{*}{scenes} & & \multicolumn{5}{c}{w.o. correspondence} & & \multicolumn{5}{c}{Ours} \\ 
\cline{3-4} \cline{6-7} \cline{9-10} \cline{12-13} 
 &  & PSNR  & SSIM & & $\text{RPE}_t$  &$\text{RPE}_r$ & & PSNR & SSIM & & $\text{RPE}_t$ & $\text{RPE}_r$ \\ \hline
Church & & 27.95 & 0.88 & & 0.031 & 0.089 & & 32.14	& 0.96 & & 0.006 & 0.016 \\ 
Barn & & 28.20 & 0.89 & & 0.127	& 0.194 & & 33.19 & 0.94 & & 0.029 & 0.030 \\ 
Museum & & 27.95 & 0.83 & & 0.074 & 0.212 & & 31.62	& 0.94 & & 0.047 & 0.203 \\ 
Family & & 29.12 & 0.83 & & 0.051 & 0.028 & & 34.80	& 0.97 & & 0.024 & 0.020 \\
Horse & & 29.43	& 0.87 & & 0.135 & 0.061 & & 35.45 & 0.97 & & 0.109	& 0.053\\
Ballroom & & 28.19 & 0.84 & & 0.056	& 0.064 & & 33.91 & 0.97 & & 0.033 & 0.020 \\ 
Francis & & 28.57 & 0.79 & & 0.103 & 0.159 & & 33.80 & 0.92 & & 0.026 & 0.147 \\ 
Ignatius & & 26.66 & 0.76 & & 0.150 & 0.044 & & 31.14 & 0.94 & & 0.027 & 0.012\\ \hline
mean & & 28.26 & 0.84 & & 0.091	& 0.106 & & \textbf{33.26} & \textbf{0.95} & & \textbf{0.037} & \textbf{0.063} \\
\bottomrule
\end{tabular}

\caption{\textbf{Ablation for Correspondence on Tanks and Temples.} Performance on both novel view synthesis and camera pose estimation. The best results of means are highlighted in bold.}

\vspace{-4mm}
\label{table:abs_grow}
\end{table}

%% file: tables/abs_colmap.tex
\begin{table}[tp]
\centering
\setlength{\tabcolsep}{3.7pt}
\small
\begin{tabular}{cccccccccc}
\toprule
\multirow{2}{*}{scenes} & & \multicolumn{3}{c}{Ours} & & \multicolumn{3}{c}{COLMAP + 3DGS} \\ 
\cline{3-5} \cline{7-9}
 &  & PSNR  & SSIM & LPIPS & & PSNR & SSIM & LPIPS \\ \hline
Church & & 32.14 & 0.96 & 0.08 & & 29.93 & 0.93 & 0.09 \\
Barn & & 33.19	& 0.94 & 0.07 & & 31.08 & 0.95 & 0.07 \\
Museum & & 31.62 & 0.94 & 0.08 & & 34.47 & 0.96 & 0.05  \\ 
Family & & 34.80 & 0.97 & 0.04 & & 27.93 & 0.92 & 0.11\\
Horse & & 35.45	& 0.97	& 0.04 & & 20.91 & 0.77 & 0.23\\
Ballroom & & 33.91 & 0.97 & 0.04 & & 34.48 & 0.96 & 0.04 \\ 
Francis & & 33.80 & 0.92 & 0.13 & & 32.64 & 0.92 & 0.15 \\ 
Ignatius & & 31.14 & 0.94 & 0.06 & & 30.20 & 0.93 & 0.08 \\ \hline
mean & & \textbf{33.26} & \textbf{0.95} & \textbf{0.07} & & 30.20 & 0.92 & 0.10\\
\bottomrule
\end{tabular}
\vspace{-2mm}
\caption{\textbf{Comparison to 3DGS trained with SfM poses}. We
report the performance of novel view synthesis using ours and vanilla 3DGS. The best results of means are highlighted in bold.}
\vspace{-3mm}
\label{table:abs_colmap}
\end{table}